# A Public Reference Implementation of the RAP Anaphora Resolution Algorithm

Long Qiu, Min-Yen Kan, Tat-Seng Chua

School of Computing
National University of Singapore
3 Science Drive 2
Singapore 117543
{qiul,kanmy,chuats}@comp.nus.edu.sg

**Abstract**

This paper describes a standalone, publicly-available implementation of the Resolution of Anaphora Procedure (RAP) given by Lappin and Leass (1994). The RAP algorithm resolves third person pronouns, lexical anaphors, and identifies pleonastic pronouns. Our implementation, JavaRAP, fills a current need in anaphora resolution research by providing a reference implementation that can be benchmarked against current algorithms. The implementation uses the standard, publicly available Charniak (2000) parser as input, and generates a list of anaphora-antecedent pairs as output. Alternately, an in-place annotation or substitution of the anaphors with their antecedents can be produced. Evaluation on the MUC-6 co-reference task shows that JavaRAP has an accuracy of 57.9%, similar to the performance given previously in the literature (e.g., Preiss 2002).

## 1. Introduction

Anaphora refers to the phenomenon where a word or phrase in a sentence is used to refer to an entity introduced earlier into the discourse, and the word or phrase is said to be an anaphor, or anaphoric. Accordingly, anaphora resolution is the process of identifying an anaphor's antecedent(s) thus to conceptually link it with its referent. Given that anaphora is resolved correctly, it can significantly augment the performance of downstream NLP applications, question answering (Vicedo and Ferrández, 2000) for instance. While there have been many approaches to anaphora resolution in the literature, the comparative evaluation of such algorithms has been hampered by a number of factors, including the lack of publicly available reference implementations. To address this problem, we created JavaRAP, a Java-based implementation of the seminal Resolution of Anaphora Procedure (RAP) algorithm (Lappin and Leass, 1994) and made it freely available. We expect JavaRAP to facilitate research along with other implementations of anaphora resolution approaches.

RAP is an algorithm for identifying both intersentential and intrasentential antecedents of third person pronouns (in their nominative, accusative or possessive case) and lexical anaphors (including *reflexives* -- pronouns like "myself", "yourself", *etc.* which are used when the complement of the verb is the same as the subject or to emphasize the subject or object, and *reciprocals* -- phrases like "each other" and "one another" showing that an action is two-way). The assumed input for RAP is the syntactic representations generated by McCord's Slot Grammar parser (1990). For lexical anaphors, an anaphor binding algorithm is applied to find the possible antecedents, while for third person pronouns, a syntactic filter rules out the noun phases that are unlikely to be the antecedents. If there remain more than one candidate, a salience measure is used and the candidate with the highest salience weight is selected.

Following RAP, JavaRAP resolves the anaphora by first extracting a list of all noun phrases in the input and a list of resolvable anaphors -- third person pronouns and lexical anaphors. Each anaphor is paired with noun phrases within a small sentence window. The resulting anaphor/antecedent-candidate pairs are then checked for agreement (gender, person and number) and filtered through the anaphor binding algorithm or syntactic filter, whichever applies. The candidate with the highest salience weight is selected as the actual antecedent.

The remainder of this paper is organized as follows. In section 2, we present in details how each step of RAP algorithm is implemented. Section 3 shows the resolution procedure of JavaRAP. What follows is a brief introduction to the associated tools. Section 5 covers the evaluation environment and the results.

## 2. RAP in Details and Its Implementation

### 2.1 Parser

In the original RAP implementation, McCord's Slot Grammar parser is used to provide the detailed parse. As parsers providing such rich output were not widely available, this was seen as a limitation. Kennedy and Boguraev (1996) addressed this by implementing a "knowledge poor" version of RAP using just a part of speech (POS) tagger as input. As the POS tagger does not annotate the chunks or provide head-argument/head-adjunct information required by RAP, Kennedy and Boguraev approximate this by a combination of phrasal grammar and text patterns.

In contrast, our JavaRAP utilizes the publicly available "knowledge rich" Charniak parser as input. Sample output generated by it is shown in Figure 1.

```
Input: <s> (``He'll work at the factory.'') </s>
Output:
(S1 (PRN (-LRB- -LRB-)
   (S (`` ``)
     (NP (PRP He))
     (VP (MD 'll)
       (VP (VB work) (PP (IN at) (NP (DT the) (NN factory)))))
     (. .)
     ('' ''))
   (-RRB- -RRB-)))
```

Figure 1: Sample Input and Output of the Charniak Parser

Head-argument/head-adjunct relations and grammatical roles, which are required by the RAP algorithm, are not given by the parser. JavaRAP recovers them by using structure information of the verb/noun phrases involved (The details are covered in section 2.2 and 2.4, respectively). As such, we believe that our implementation is closer in fidelity to the original algorithm.

## 2.2 Syntactic Filter and Anaphor Binding Algorithm

Before we move to the syntactic filter and anaphor binding algorithm of RAP, it is necessary to introduce the terminology that appears in their description (and in introduction to salience factors in section 2.4). Lappin and Leass give the following definitions:

1. "… a phrase P is in the *argument domain* of a phrase N iff P and N are both arguments of the same head.
2. … P is in the *adjunct domain* of N iff N is an argument of a head H, P is the object of a preposition PREP, and PREP is an adjunct of H.
3. P is in the *NP domain* of N iff N is the determiner of a noun Q and (i) P is an argument of Q, or (ii) P is the object of a preposition PREP and PREP is an adjunct of Q.
4. A phrase P is *contained* in a phrase Q iff 1) P is either an argument or an adjunct of Q, i.e. P is *immediately contained* in Q, or 2) P is immediately contained in some phrase R, and R is contained in Q."

According to RAP, a third person pronoun *P* is not coreferential with a (non-reflexive or non-reciprocal) noun phrase *N* within a sentence if one of the following rules holds (*syntactic filter*):

1. *P* and *N* have incompatible *agreement features* (number, people and gender): ("The woman said that he is funny.");
2. *P* is in the argument domain of *N* ("She likes her.");
3. *P* is in the adjunct domain of *N* ("She sat near her.");
4. *P* is an argument of a head *H*, *N* is not a pronoun, and *N* is contained in *H* ("He believes that the man is amusing.");
5. *P* is in the NP domain of *N* ("John's portrait of him is interesting."); and
6. *P* is a determiner of a noun *Q*, and *N* is contained in *Q* ("His portrait of John is interesting.").

For lexical anaphors, a noun phrase *N* is a possible antecedent of a lexical anaphor *A* if their agreement features are compatible and one of the following rules holds (*anaphor binding algorithm*):

1. *A* is in the argument domain of *N*, and *N* fills a higher argument slot than *A* ("They wanted to see themselves.");
2. *A* is in the adjunct domain of *N* ("He worked by himself.");
3. *A* is in the NP domain of *N* ("John likes Bill's portrait of himself.";
4. *N* is an argument of a verb *V*. Meanwhile, there is a noun phrase *Q* in the argument domain or the adjunct domain of *N* such that *Q* has no noun determiner, and (i) *A* is an argument of *Q*, or (ii) *A* is an argument of a preposition *PREP* which is an adjunct of *Q* ("They told stories about themselves."); and
5. *A* is a determiner of a noun *Q*, and (i) *Q* is in the argument domain of *N* and *N* fills a higher argument slot than *Q*, or (ii) *Q* is in the adjunct domain of *N* ("John and Marry like each other's portraits.").

To find the agreement features of pronouns is straightforward as they are reflected in the pronouns themselves. The problem is more complicated for other noun phrases. In JavaRAP, the agreement features of these noun phrases are obtained as follows:

*Number*: This feature is set as 'true' for singular noun phrases and 'false' for plural ones. If a noun phrase is the agent of a verb phrase, we attempt to decide its number by inspecting the tag of the verb phrase. Otherwise, the tag of the noun phrase itself is checked. If the noun phrase contains more than one word, the existence of the word 'and' or the tag of the phrase's head either can serve as the clue whether the phrase's number is singular or plural. This feature remains 'unknown' if all these methods fail.

*People and Gender*: For other singular and plural noun phrases, their default people feature is "third". However, two special cases are considered. For a plural noun phrase, the people feature is set as "first" if it contains a first person pronoun in its nominative or accusative case. If the noun phrase contains not a first but a second person pronoun in its nominative/accusative case, the feature is set as "second". Two Christian first name lists (male and female) available from U.S Census Bureau's website (www.census.gov/genealogy/names/) are used to detect the gender feature of noun phrases. Once the string of the noun phrase is found in one of these lists, its gender feature is set accordingly. Otherwise, it remains "unknown": there is no default value for gender. A successful looking-up also generates *animacity* information of the noun phrase and that is used as an auxiliary agreement feature.

The agreement features' compatibility of a pronoun and a noun phrase is examined by a *morphological filter*. It declares two noun phrases are non-matching in their agreement features only if at least one feature explicitly disagrees. The value "unknown" is regarded to agree with any value of the feature.

As to *head-argument* and *head-adjunct* relationship, *etc.*, as mentioned earlier, the required information is obtained by inspecting the parse tree structure:

1. An NP is in the argument domain of another NP if one is a child of the following sibling VP of the other, or the two NPs are connected by a conjunction and they together form a sibling of a VP. The VP is the argument head of both NPs;
2. An NP is in the adjunct domain of another NP if the former is a child of a PP, which is in turn a child of a VP, and the latter is either a child or a sibling of the VP. The VP is the adjunct head of the former NP;
3. An NP is in the NP domain of another NP if the former NP is a child of a PP; the PP has a proceeding sibling NP, the children of which include the later NP and a following POS;
4. An NP is contained in a VP if the VP is the NP's argument head or adjunct head; an NP is contained in another NP if the former is a child of the latter's sibling PP. Furthermore, an NP is considered to be contained in a VP/NP if it is contained in a phrase Q and Q is contained in the VP/NP.

### 2.3 Pleonastic Pronouns

RAP also identifies pleonastic pronouns, the pronouns that have no referent. The pronoun "it" is commonly used as the pleonastic pronoun in English. Typically it appears with a *modal adjective* ("…it is important to…") or a *cognitive verb* in its passive participle form ("…it is recommended that..."), *etc*. RAP uses a modal adjective list and a cognitive verb list to detect pleonastic pronouns appearing in the following patterns:

1. it is *Modaladj* that *S*,
2. it is *Modaladj* (for *NP*) to *VP*,
3. it is *Cogv-ed* that *S*,
4. it seems / appears / means / follows (that) *S*,
5. *NP* makes / finds it *Modaladj* (for *NP*) to *VP*,
6. it is time to *VP*, and
7. it is thanks to *NP* that *S*,

where *Modaladj* stands for a modal adjective and *Cogv-ed* stands for the passive participle of a cognitive verb.

JavaRAP performs pattern matching for each pronoun 'it' and it is labeled as pleonastic if the matching is successful. Syntactic variants of these patterns (it is not /may be *Modaladj* that…, wouldn't it be *Modaladj*…, *etc*.) are considered also, as suggested by RAP. Worth pointing out here is that the modal adjective and cognitive verb lists used in JavaRAP are the same as RAP uses. They could be further augmented in order to improve the recall in identifying pleonastic pronouns.

### 2.4 Salience Factors

Each candidate antecedent of an anaphor has an associated salience weight computed from a set of salience factors. Table 1 shows all the salience factors and their initial weights. With the exception of the sentence recency factor (the weight of which is non-zero only if the candidate is in the same sentence as the anaphor is), the weights of all other factors degrade in half each time the number of sentences between the candidate and the anaphor increases.

| Factor | Initial Weight |
|---|---|
| Sentence Recency | 100 |
| Subject Emphasis | 80 |
| Existential Emphasis | 70 |
| Accusative Emphasis | 50 |
| Indirect Object and Oblique Complement Emphasis | 40 |
| Head Noun Emphasis | 80 |
| Non-adverbial Emphasis | 50 |

Table 1: Salience factors and their initial weights

The Charniak parser does not show the required grammatical information to calculate salience weights. To tackle this problem, JavaRAP extracts the information from the parse tree structure by using the following rules:

1. An NP is a *subject* if its parent is S;
2. An NP is *existential* if it is the second (from the left) child of a VP and the proceeding sibling of the VP is an NP whose first child is EX;
3. An NP is *direct object* if it is the only NP child of a VP, or the second NP child of a VP, while in the latter case the first NP child is an indirect object;
4. An NP is a *head noun* if it is not contained in another NP;
5. An NP is not contained in an *adverbial prepositional phrase* if there is no ADVP among its ancestors.

JavaRAP takes the salience weight for each factor as they are proposed in RAP.

### 2.5 Equivalence Class

In RAP, noun phrases identified to be in the same "anaphoric chain" form an equivalence class of discourse referents. The salience weight associated with this class is the sum of the weights of all the salience factors that present in the group.

In JavaRAP, each noun phrase has a set of salience factors associated with it locally. The salience weight of it is computed online during the resolution process by a simple summation, and followed by degradation, if applicable. The idea of equivalence class is realized by keeping merging the salience factors of the last two noun phrases in the anaphoric chain. In this way, it is guaranteed that the latest noun phrase has all the salience factors that its ancestors have.

## 3. Resolution Procedure

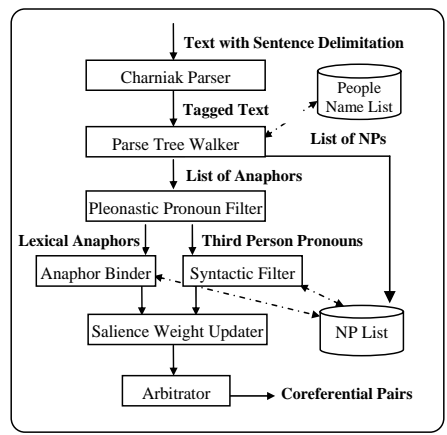

Figure 2: JavaRAP System Structure

Figure 2 shows the structure of JavaRAP (the Charniak parser included). The implemented resolution procedure is as the following:

1. The Charniak parser takes text with sentence delimitation as input and generates a parse tree;
2. The Parse Tree Walker extracts two lists. One contains all the noun phrases in the text and the other, the third person pronouns and reflexive pronouns. Their agreement features, head-argument/head-adjunct information, whether they are contained in other phrases sand all the salience factors except sentence recency are annotated during extraction. The pleonastic pronouns identifiable are also labeled;
3. Each anaphor forms a pair with each item in a subset of the noun phrases (currently JavaRAP only considers noun phrases contained within three sentences proceeding the anaphor and those in the sentence where the anaphor resides). These antecedent candidate-anaphor pairs are examined by the anaphor binding algorithm or the syntactic filter, depending on whether the anaphor is lexical or a third person pronoun. Noun phrases unlikely to be antecedents are removed;
4. Remaining antecedent candidates are ranked by their salience weights and the top one is proposed as the actual antecedent, the one closer to the anaphor is favored in case of a tie.

## 4. Associated Tools

We have packaged two utilities along with JavaRAP to enable end-to-end anaphora resolution. As the Charniak parser expects sentence boundaries to be marked, we provide an efficient, configurable rule-based sentence splitter that handles multiple input formats. It works by checking each instance of sentence-ending punctuations (period, question mark, exclamation mark, quotation mark) and deciding whether to delimit the sentence there. Following limitations are noticed:
1. New line feeds are not considered as potential boundaries. Therefore, titles/subtitles of articles are always appended to the following sentences;
2. A list of abbreviations like "Mr.", "Mt.", *etc.* are used to filter out false candidate boundaries. However, two sentences are mistakenly concatenated if the leading one does contain such an abbreviation in the very end;
3. The sentence splitter is case-sensitive. The accuracy of it will drop if the article contains only capitalized letters and it will not be able to delimit sentences if they are all in lower case.

The sentence splitter is independent of the resolver and has been used for other NLP applications.

To facilitate resolver evaluation, we also include a tool to perform pair-wise comparison between a gold standard annotated text and resolver output. Annotations are accepted in the standard MUC-6 co-reference annotation convention and equivalence classes could be restored based on them. Before this comparator can work correctly, all the instances of third person pronouns have to be annotated in the gold standard text.

## 5. Evaluation

As an evaluation, we chose to use the training set for MUC-6's co-reference task (to identify co-reference relations amongst noun phrases) as the test set because of its comprehensive annotations. In particular, co-reference relations for third-person pronouns and third-person reflexive pronouns are annotated. Out of the total 235 lexical anaphors and third person pronouns annotated in the test set, JavaRAP labels 136 correctly. That gives an accuracy of 57.9%. It is comparable to the performance of the RAP implementation mentioned in (Preiss, 2002), as shown in Table 2. Preiss' implementation uses sentences from the BNC (British National Corpus) as the test set while we use MUC-6 data. The difference between these two test sets may result in different performance of the Charniak parser in term of accuracy. Furthermore, both implementations extract the grammatical roles by applying certain hand-crafted rules on the parse tree. The dissimilarity of the grammatical role extractors is another factor that could make the overall performance of the two implementations different.

|  | Parser | Test Set | Accuracy |
|---|---|---|---|
| **JavaRAP** | Charniak | MUC-6 | 57.9% |
| **Preiss'** | Charniak | BNC | 61% |

Table 2: JavaRAP and Preiss' implementation of RAP

## 6. Conclusions

We present JavaRAP, a platform-portable, standalone implementation of the classic Lappin and Leass anaphora resolution algorithm. It can be freely downloaded from our website (www.comp.nus.edu.sg/~qiul/NLPTools) with the utilities (as noted in section 4) to enable end-to-end resolution and evaluation. For progress in anaphora resolution research to occur, Mitkov (2000) argues for greater transparency and sharing of corpora and resolvers. We view our work as a step in this direction. There are some simplifications and approximations made in the implementation process so that it is possible to build it in a short period of time and with few NLP resources. On the MUC-6 co-reference task JavaRAP's performance is comparable to similar, proprietary implementations. We hope that researchers will use this implementation as a reference point for future comparative evaluations on different corpora. For future work, we plan to provide manually-corrected, perfect syntactic and semantic information to the algorithm to benchmark its upper bound performance.